\documentclass[10pt, a4paper, onecolumn]{article}

\usepackage[numbers]{natbib}

\usepackage[T1]{fontenc}
\usepackage{lmodern}

\usepackage{natbib}
\usepackage[utf8]{inputenc} 
\usepackage{booktabs}       
\usepackage{amsfonts}       
\usepackage{nicefrac}       
\usepackage{microtype}      
\usepackage{xcolor}         
\usepackage{graphicx}
\usepackage{amsmath,amssymb,amsfonts}
\usepackage{bm}
\usepackage{url}

\usepackage{array,multirow,graphicx}

\usepackage{geometry}
\newgeometry{vmargin={15mm}, hmargin={12mm,17mm}} 

\title{Technical Report:\\ Combining knowledge from Transfer Learning during training and Wide Resnets}

\author{
	Wolfgang Fuhl \\
	Mühringen, 72160 \\
	\texttt{wolfgang-fuhl@gmx.de} \\
}

\begin{document}
	
	\maketitle
	
	\begin{abstract}
		In this report, we combine the idea of Wide ResNets and transfer learning to optimize the architecture of deep neural networks. The first improvement of the architecture is the use of all layers as information source for the last layer. This idea comes from transfer learning, which uses networks pre-trained on other data and extracts different levels of the network as input for the new task. The second improvement is the use of deeper layers instead of deeper sequences of blocks. This idea comes from Wide ResNets. Using both optimizations, both high data augmentation and standard data augmentation can produce better results for different models.
		Link: \url{https://github.com/wolfgangfuhl/PublicationStuff/tree/master/TechnicalReport1/Supp}
	\end{abstract}

	\section{Introduction}

	DNNs~\cite{fuhl2021tensor} have found their way into a variety of fields~\cite{WF042019,UMUAI2020FUHL}. In eye tracking~\cite{fuhl2022groupgazer,fuhl2022pistol}, they are already used for scanpath analysis~\cite{fuhl2022hpcgen,C2019,FFAO2019}, as well as other approaches based on machine learning~\cite{fuhl2018simarxiv,ICMIW2019FuhlW1,ICMIW2019FuhlW2,EPIC2018FuhlW}, feature extraction~\cite{fuhl2022pistol} such as pupil~\cite{fuhl20211000,WTCKWE092015,WTTE032016,062016,CORR2017FuhlW2,CORR2016FuhlW,WDTE092016,WTCDAHKSE122016,WTCDOWE052017,WDTTWE062018,VECETRA2020,CORR2017FuhlW1,ETRA2018FuhlW}, iris~\cite{ICCVW2019FuhlW,CAIP2019FuhlW,ICCVW2018FuhlW} and eyelids~\cite{WTDTWE092016,WTDTE022017,WTE032017}, eyeball estimation~\cite{NNETRA2020}, and also for eye movement classification~\cite{FCDGR2020FUHLARX,FCDGR2020FUHL,MEMD2021FUHL}. In recent years, there have been a number of new large data sets~\cite{fuhl2021teyed} that have also been annotated using modern machine learning approaches. Another important aspect is the anonymisation of the data~\cite{RLDIFFPRIV2020FUHL,GMS2021FUHL}, in which more and more research is done. Other areas include human computer interaction, medicine, robotics, industrial sensor evaluation, and many more.
	
	A lot of research is also being done in the field of visualization~\cite{ROIGA2018,ASAOIB2015,AGAS2018}. Here one tries to illustrate the data with different statistics and distributions. The visualizations also provide a way to better understand deep neural networks and other machine learning methods. In the field of data analysis, machine learning is often used to visualize values as functions and to reduce high-dimensional feature spaces. Based on the undestanding, it is also important to validate models~\cite{ICMV2019FuhlW,NNVALID2020FUHL,NNVALID2020FUHLARX}. 
	
	With all the applications and improvements achieved by machine learning and in particular by deep neural networks, this research field has established itself as state of the art. Research in the field of machine learning is also continuously delivering new improvements such as tree convolution~\cite{AAAIFuhlW}, neural pooling~\cite{NNPOOL2020FUHL}, rotating convolution operators~\cite{RINGRAD2020FUHL} and many more~\cite{NORM2020FUHL}. In this work, we describe the combination of two existing ideas that allow to obtain better generalized networks and to reduce the loss of information due to spatial reduction. Together with the idea of wide ResNets, parameters and operations in networks can be saved.
	
	\pagebreak
	
	\section{Method}
	\begin{figure}[h]
		\centering
		\includegraphics[width=0.95\textwidth]{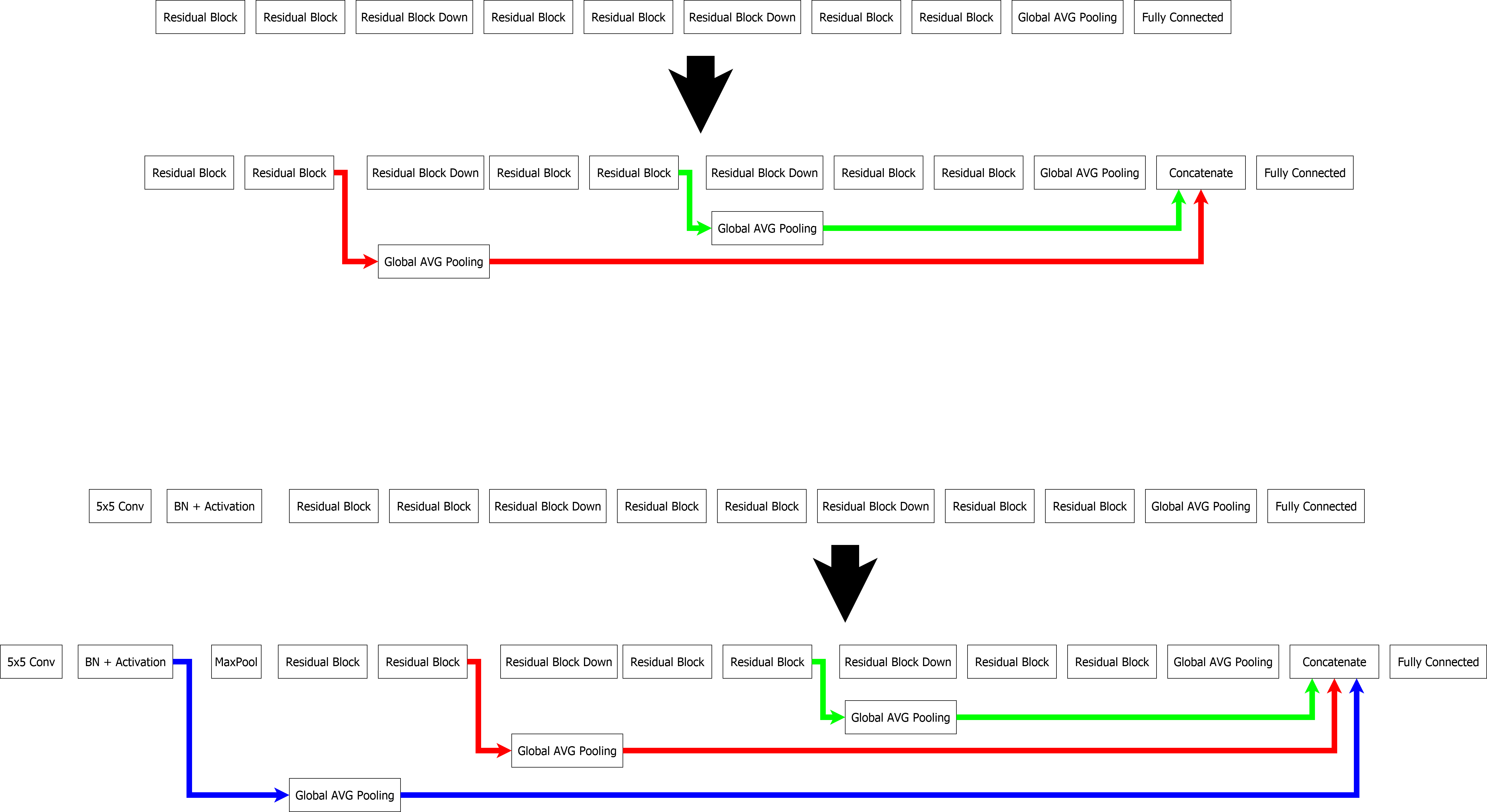}
		\caption{Shows the basic procedure for optimizing an architecture with the presented method. Before each residual block with a spatial reduction, an additional connection is added. This connection performs an average pooling along the depth of the layer and provides the result to the last layer as an additional source of information. The residual nets are shown here only partially since to reduce the size of the image.}
		\label{fig:workflow}
	\end{figure}

\begin{figure}[h]
	\centering
	\includegraphics[width=0.95\textwidth]{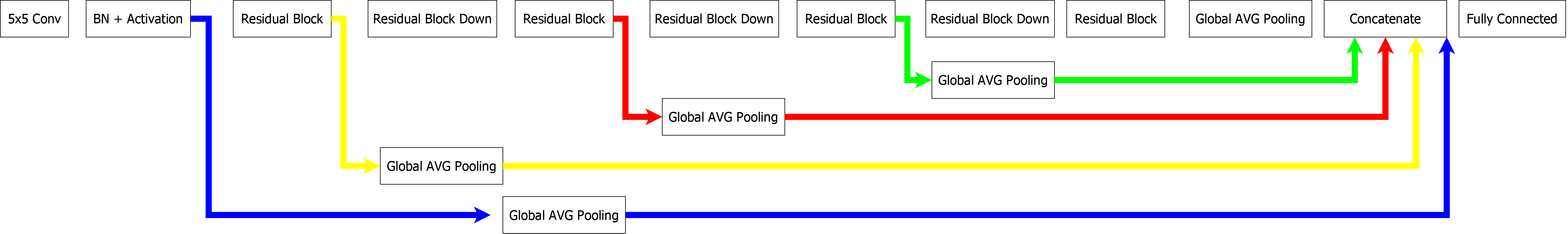}
	\caption{The proposed architecture. Each residual stage has double the depth of the standard ResNet-32 and the propagation to the last fully connected layer in combination with the depthwise pooling. Additionally, we have a convolution + BN + activation at the beginning, as it is usually done for large image inputs like image net.}
	\label{fig:newnet}
\end{figure}

In Figure~\ref{fig:workflow} it is shown how to transform nets with residual~\cite{he2016deep} or maximum propagation~\cite{fuhl2021maximum} connections to the proposed architecture. This can be applied for all networks and steams from the idea of using multiple layers of a feature extraction network.

In Figure~\ref{fig:newnet} a new architecure is shown. It has one additional convolution in the beginning (Similar to resnets for larger inputs like image net). Compared to ResNets all levels are decreased to only contain two residual blocks. Additionally, each level has a depthwise pooling and forward propagation integrated.
	
	\pagebreak
	
	\section{Evaluation}
	
	\begin{table}[htb]
		\caption{The results of the proposed approch to standard models on Cifar 100~\cite{krizhevsky2009learning}. We used the standard train and validation split as well as we used the same resolution of 32 by 32. We also used only one GPU (RTX 3050 ti 8GB RAM).\\
		Shift + Flip: Cropping a image region (32 by 32) out of the original image with a possible shift of 4 in each direction. Pixels outside the original image are filled with zeros. Flipping was done left right only. We used the standard scheduler reduction of 0.1 after each 100 epochs, an initial learning rate of 0.01 and a batch size of 10 with the SGD and Momentum optimizer. Each net was trained for 400 epochs.\\
		For WideResNet~\cite{zagoruyko2016wide} and FastAutoAugment~\cite{lim2019fast} we used the implementation provided here https://github.com/kakaobrain/fast-autoaugment. They use the sinus scheduler and many additional data augmentations. We had to reduce the batch size to half due to the available RAM on the GPU.
		}
		\label{tbl:evalFake}
		\centering
		\begin{tabular}{llc}
			\toprule
			Model & Augmentation & Cifar 100 Accuracy \\ \hline
			ResNet32 & Shift + Flip& 76.7 \\ 
			ResNet32 + Proposed & Shift + Flip& 77.6 \\ 
			ResNet32 (Double depth) & Shift + Flip& 76.9 \\ 
			ResNet32 (Double depth) + Proposed & Shift + Flip& 78.11 \\ 
			NewNet (Figure~\ref{fig:newnet})  & Shift + Flip& 78.94\\ 
			NewNet (Figure~\ref{fig:newnet}, double depth)  & Shift + Flip& 79.67 \\ 
			WideResNet 28x10 & FastAutoAugment & 82.01 \\ 
			WideResNet 28x10 + Proposed & FastAutoAugment & 83.52 \\ \hline
		\end{tabular}
	\end{table}

	\section{Acknowledgements}
	I know that evaluations on multiple data sets as well as more models are neccessary to show the improvement. But since I have to pay for the energy and I am not employed at a research department, I leave this to others if they are interested.
	
	\bibliographystyle{plain}
	\bibliography{template}

\end{document}